\documentclass{article}
\usepackage{ijcai25}

\usepackage{times}
\usepackage{soul}
\usepackage{url}
\usepackage[hidelinks]{hyperref}
\usepackage[utf8]{inputenc}
\usepackage[small]{caption}
\usepackage{graphicx}
\usepackage{amsmath}
\usepackage{amsthm}
\usepackage{booktabs}
\usepackage{algorithm}
\usepackage{algorithmic}
\usepackage[switch]{lineno}

\usepackage{amssymb}
\usepackage{multirow}
\usepackage{subfigure}
\usepackage{booktabs} 

\title{AHSG: Adversarial Attack on High-level \\Semantics in Graph Neural Networks}

\author{
Kai Yuan$^1$
\and
Jiahao Zhang$^1$\and
Yidi Wang$^2$\and
Xiaobing Pei$^1$\\
\affiliations
$^1$Huazhong University of Science and Technology\\
$^2$Great Bay University\\
\emails
m202376878@hust.edu.cn,
m202476982@hust.edu.cn,
w\_yidi@foxmail.com,
xiaobingp@hust.edu.cn
}

\begin{document}

\maketitle

\begin{abstract}
Adversarial attacks on Graph Neural Networks aim to perturb the performance of the learner by carefully modifying the graph topology and node attributes. Existing methods achieve attack stealthiness by constraining the modification budget and differences in graph properties. However, these methods typically disrupt task-relevant primary semantics directly, which results in low defensibility and detectability of the attack. In this paper, we propose an Adversarial Attack on High-level Semantics for Graph Neural Networks (AHSG), which is a graph structure attack model that ensures the retention of primary semantics. By combining latent representations with shared primary semantics, our model retains detectable attributes and relational patterns of the original graph while leveraging more subtle changes to carry out the attack. Then we use the Projected Gradient Descent algorithm to map the latent representations with attack effects to the adversarial graph. Through experiments on robust graph deep learning models equipped with defense strategies, we demonstrate that AHSG outperforms other state-of-the-art methods in attack effectiveness. Additionally, using Contextual Stochastic Block Models to detect the attacked graph further validates that our method preserves the primary semantics of the graph.
\end{abstract}

\section{Introduction}

GNNs \cite{1,2,3}have achieved significant success in graph representation learning. Due to their excellent performance, GNNs have been applied to various analytical tasks, including node classification \cite{11,12}, link prediction \cite{13}, and graph classification \cite{14}. Recent studies have shown that similar to traditional deep neural networks, GNNs suffer from poor robustness when facing specially designed adversarial attacks. Attackers can generate graph adversarial perturbations to deceive GNNs by manipulating graph structures and node features \cite{16,17,18,19,20,24}, or generating new nodes and adding them to the original graph \cite{22,23}. By understanding the ways in which models are vulnerable to attacks, a range of defence methods are designed that perform better when confronted with adversarial samples \cite{26,27,28,29,30}.

\begin{figure}[t]
\vskip -0.1in
\centering
\includegraphics[width=0.47\textwidth]{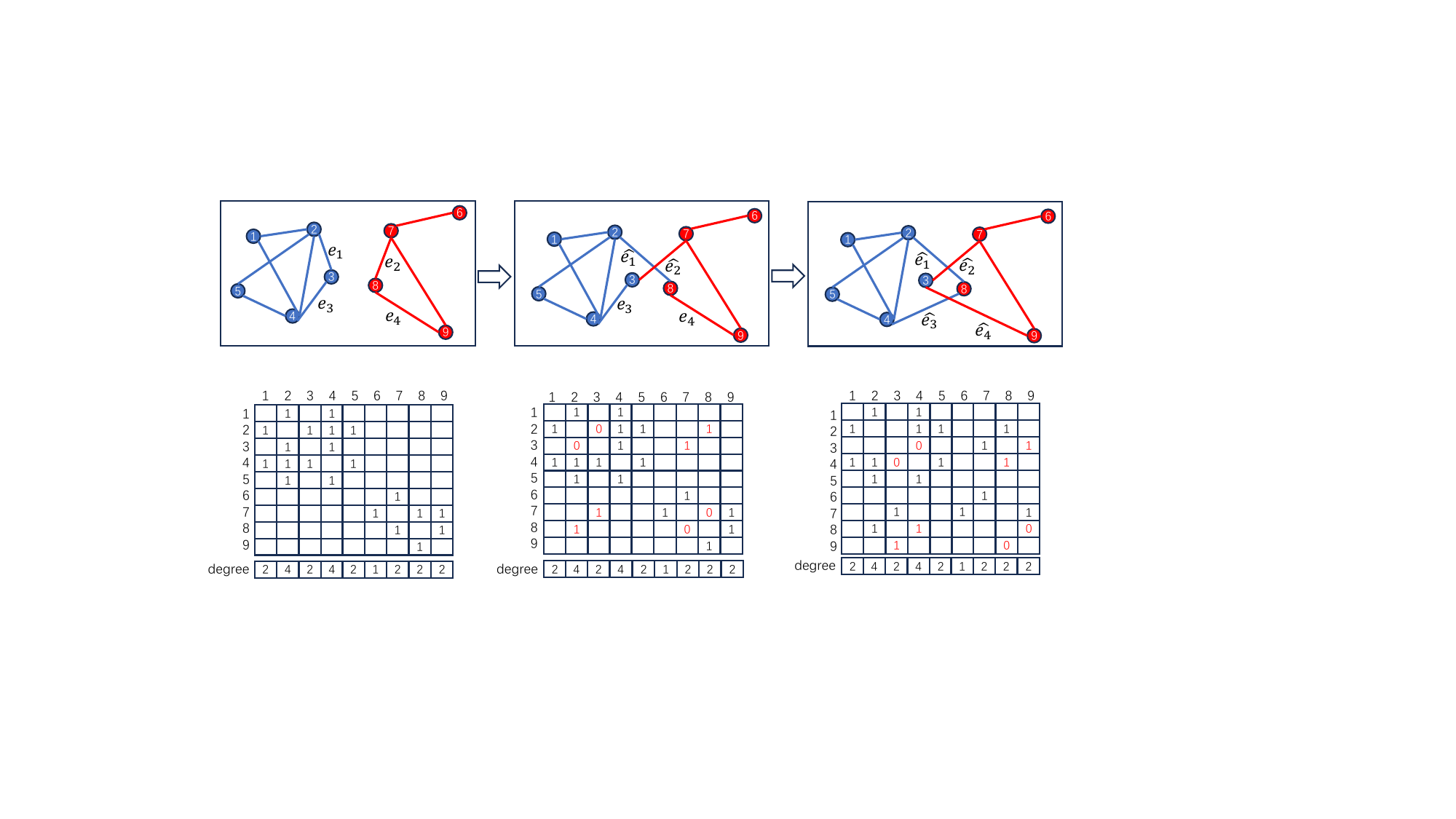} 
\vskip -0.1in
\caption{
The primary semantics of graph are disrupted under the constraints of degree distribution.}
\label{fig1}
\vskip -0.15in
\end{figure}

A core principle of attacking neural networks is that the attacker should preserve the primary semantics of the original data after adding perturbations. This semantic invariance ensures the practical significance of the attack. To achieve this, adversarial attacks usually require the attacker to make only minimal perturbations. Therefore, most existing research on graph adversarial attacks \cite{16,17,18,19,20} restricts attackers to modifying a limited number of edges or nodes. However, it remains uncertain whether such attack models can effectively preserve the primary semantics. For example, real-world graphs typically have many low-degree nodes, and a small attack budget can completely remove their original neighbors. Once the edges of low-degree nodes are disrupted, their semantics are often lost. Therefore, preserving the primary semantics requires more than just managing the attack budget.

To further constrain semantic changes, existing methods introduce metrics beyond attack budget limitations. \cite{19,23} suggest using different global graph properties as proxies for semantics, such as degree distribution and homophily. However, even under such constraints, the primary semantics can still be disrupted. Figure \ref{fig1} illustrates this phenomenon with a binary classification task. Assume we select two edges \( e_1 = (i,j) \) and \( e_2 = (u,v) \) from the edge set of an undirected graph, and replace them with \(\widehat{e_1} = (i,v)\) and \(\widehat{e_2} = (u,j)\). Since \(\deg(i)\), \(\deg(j)\), \(\deg(u)\), and \(\deg(v)\) remain unchanged, this process preserves the degree distribution of the graph. We can continue to modify the edges according to this rule until the attack budget is exhausted. Finally, we successfully modifies the semantics of nodes 3 and 8 while maintaining the overall graph distribution, demonstrating the infeasibility of method \cite{19}. Therefore, neither the attack budget nor the additional constraints on graph properties can effectively preserve the primary semantics of the graph. Current research lacks a fundamental constraint aimed at maintaining primary semantics. 

To address the above question, we propose a method that generates attack graph preserving primary semantics within limited attack budget.
First, we need to find tools that represent semantics. Existing attack methods \cite{16,17,18,19,20,24} often treat graph neural networks as black boxes, focusing only on the input and output, while ignoring the rich semantic information contained in the hidden layers. The semantics of the graph are considered to be contained in the latent representations \cite{56}, so we control the change of the latent representations to control the semantic change of the attacked graph. Although the hidden layer representations in common tasks often encapsulate the primary semantics of the analyzed objects, task-irrelevant secondary semantics are not completely filtered out. These secondary semantics may play an inappropriately role in the model predictions. In this context, the construction of adversarial examples should lead the model to misjudge through exploring and exploiting the secondary semantics while keeping the primary semantics unchanged.

Subsequently, how can we get latent representations where primary semantics remain unchanged while secondary semantics change? Inspired by \cite{54}, the entities within the same category share similar primary semantics while exhibiting diverse secondary semantics. Therefore, their linearly combined representations can preserve the majority of the consensus primary semantics, while the deviations in secondary semantics can serve as the basis for crafting attacks. We use the latent representations after linear combination as input to the deeper network layers, then use gradient ascent to find latent representations with attack effects. To generate adversarial samples, a first-order optimization algorithm is employed to map the perturbed latent representations with attack effects to the attacked graph. Finally, this secondary-semantics-targeted attack method achieves lower defensibility and detectability.

Our main contributions are as follows:
\begin{itemize}
\item We propose a novel adversarial attack framework AHSG based on the graph semantics, which utilizes the class-shared latent representations with similar primary semantics as the perturbation tools.
\item We introduce a technique for reconstructing adversarial samples from perturbed representations by extracting deviations in secondary semantics from the linear combination of class-shared representations, while preserving the primary semantics.
\item We validate AHSG on multiple acknowledged benchmark datasets, demonstrating substantial improvements under different attack settings. Semantic detection experiments confirm that AHSG preserves the primary semantics of graph data.
\end{itemize}

\section{Related Work}
\subsection{Adversarial Attack on GNNs}
In graph attacks, Nettack \cite{19} introduces the first adversarial attack method on graphs. Meta-Self \cite{18} treats the input graph as a hyperparameter to be learned and modifies one edge per iteration. PGD \cite{17} overcomes the challenge of attacking discrete graph structure data. \cite{16} employ reinforcement learning, gradient-based greedy algorithms, and genetic algorithms to attack GNNs in various scenarios. EpoAtk \cite{55} improves the effectiveness of the attack by bypassing potential misinformation from the maximal gradient. GraD \cite{20} generates unweighted gradients on the graph structure, unaffected by node confidence, fully utilizing the attack budget. However, the issue of altering graph semantics during an attack remains underexplored. Although the attack budget is minimal relative to the entire graph, it can still lead to significant changes in the semantics of the attacked graph. Some works, such as Nettack \cite{19} and HAO \cite{23}, maintain certain graph properties to achieve imperceptibility. However, these constraints are necessary but not sufficient to preserve semantics.

\subsection{Adversarial Defense on GNNs}
Some studies develop corresponding defense strategies based on the characteristics of the attacks. Jaccard \cite{27} calculates the Jaccard similarity of connected node pairs and retains only those links with high similarity. Svd \cite{30} utilizes low-order approximations of the graph to enhance the performance of GCN against adversarial attacks. ProGNN \cite{28} jointly learns the graph structure and a robust GNN model from the perturbed graph. The SimPGCN \cite{29} framework enhances GCN robustness by effectively preserving node similarity. RGCN \cite{26} proposes that latent representations based on Gaussian distributions can effectively absorb the impact of adversarial attacks. In the experiment, we use AHSG along with baseline methods to attack the aforementioned defense models.

\section{Preliminary}
Given an undirected attributed graph \( G = (A, X) \) with \( n \) nodes, where \( A \in \{0,1\}^{n \times n} \) represents the adjacency matrix, and \( X \in \mathbb{R}^{n \times d} \) represents the node feature matrix. Here, \( d \) represents the feature dimension, and \( n \) represents the total number of nodes. We focus on an undirected attributed graph in this work. Formally, we denote the set of nodes as \( V = \{v_i\} \) and the set of edges as \( E \subseteq V \times V \). Each node \( v_i \) is associated with a corresponding node label \( y_i \in Y = \{0,1,\cdots,c-1\} \), where \( c \) is the total number of labels.

\subsection{Graph Neural Networks}
GNNs are specifically designed to handle graph-structured data. They leverage the message passing mechanism to aggregate information from nodes and their neighbors, thereby learning feature representations for nodes, edges, and the graph as a whole. The formula for message passing can be described as follows:
\begin{equation}
h_v^{(l)} = \text{UD}^{(l)} \left( h_v^{(l-1)}, h_t \right).
\label{eq1}
\end{equation}
The representation of node \( v \) at the \( l \)-th layer, denoted as \( h_v^{(l)} \), is computed by applying an update function \( \text{UD}^{(l)} \) to its previous representation \( h_v^{(l-1)} \) and the aggregated information \( h_t \) from its neighbors. 
\begin{equation}
h_t = \text{AG}^{(l)} \left( \left\{ \text{MG}^{(l)} \left( h_u^{(l-1)}, h_v^{(l-1)} \right) \mid u \in \mathcal{N}(v) \right\} \right).
\label{eq2}
\end{equation}
The aggregation function \( \text{AG}^{(l)} \) collects messages generated by the message function \( \text{MG}^{(l)} \), which computes the message from each neighboring node \( u \) to node \( v \) based on their respective representations \( h_u^{(l-1)} \) and \( h_v^{(l-1)} \). Here, \( \mathcal{N}(v) \) represents the set of neighboring nodes of \( v \). The initial node representation \( h^{(0)} \) is set to the node features \( X \). The learned representations are used as inputs for downstream tasks.

\subsection{Graph Adversarial Attack}
Graph adversarial attack aims to make graph neural networks produce incorrect predictions. It can be described by the following formulation:
\begin{equation}
\begin{aligned} 
& \max_{\hat{G} \in \mathrm{\Phi}(G)} \mathcal{L}\left(f_{\theta^\ast}(\hat{G}(\hat{A}, \hat{X})), Y\right)\\ 
& \text{s.t.}  \theta^\ast = \arg \min_{\theta} \mathcal{L}\left(f_{\theta}(G'(A', X')), Y\right).
\end{aligned}
\label{eq3}
\end{equation}
Here, \( f \) can be any learning task function on the graph, such as node-level embedding, node-level classification, link prediction, graph-level embedding, or graph-level classification. In this paper, we primarily focus on node-level classification. \( \mathrm{\Phi}(G) \) represents the perturbation space on the original graph \( G \). The distance between the adversarial graph and the original graph is typically measured using the attack budget or other properties. The graph \( \hat{G}(\hat{A}, \hat{X}) \) represents the adversarial sample.

When \( G' \) is equal to \( \hat{G} \), Equation \eqref{eq3} represents poisoning attack. Poisoning attack occurs during model training. It attempts to influence the model's performance by injecting adversarial samples into the training dataset. On the other hand, when \( G' \) is the unmodified original \( G \), Equation \eqref{eq3} represents evasion attack. Evasion attack occurs after the model is trained, meaning that the model's parameters are fixed when the attacker executes the attack. The attacker aims to generate adversarial samples specifically for the trained model. AHSG attacks the target model after its training, which categorizes it as an evasion attack.

\begin{figure}[t]
\centering
\includegraphics[width=0.47\textwidth]{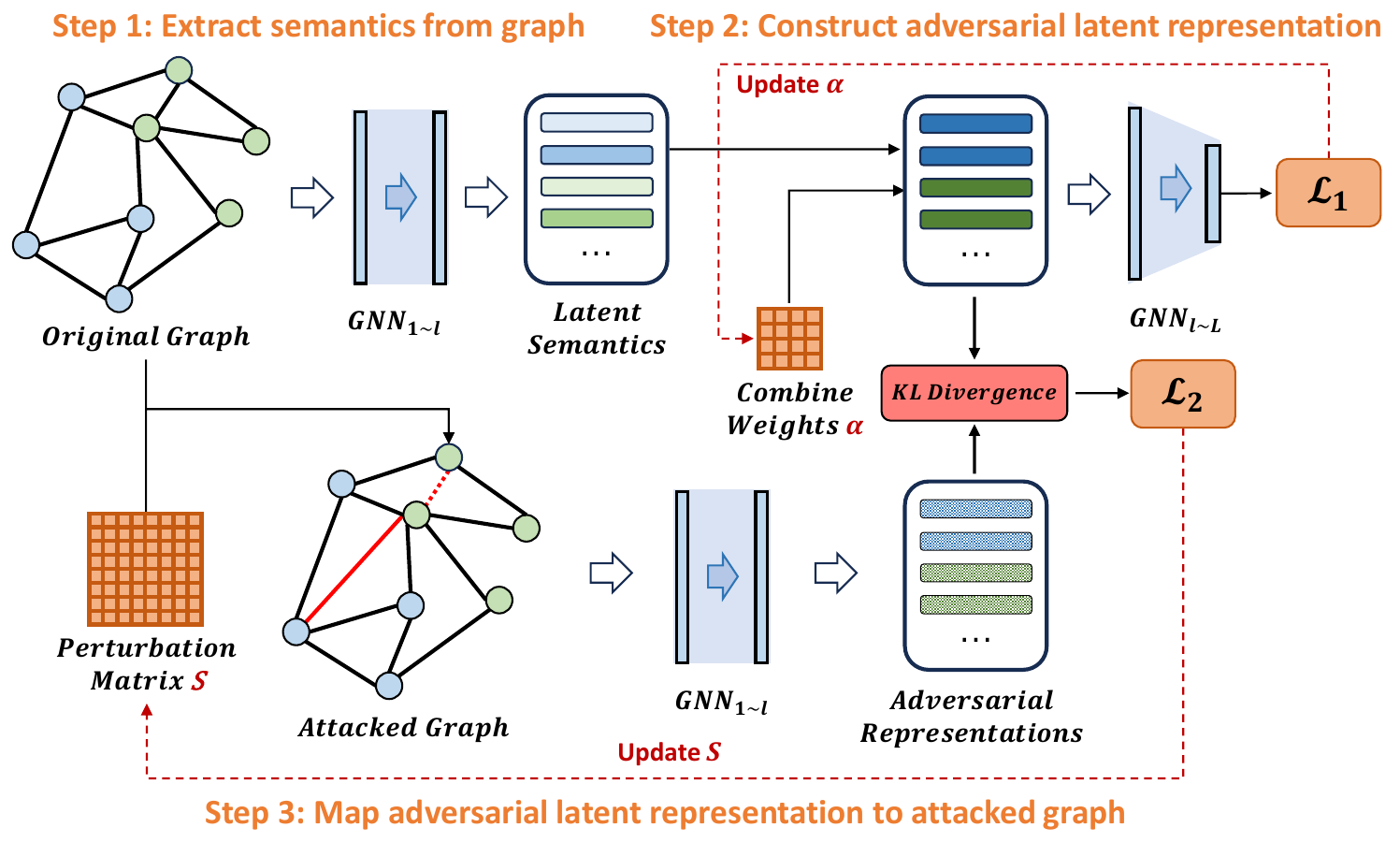} 
\caption{The framework of AHSG. It consists of three steps: extracting semantics from graph, constructing adversarial latent representations, and mapping adversarial latent representation to attacked graph. }
\label{fig2}
\end{figure}

\section{Methodology}
In this part, we first derive the formulation of the proposed model and then present its optimization algorithm.
\subsection{Architecture of AHSG}
As shown in Figure \ref{fig2}, AHSG can be divided into three stages.
\paragraph{1) Extract semantics from graph:}
The hidden layers of an unperturbed surrogate GNN encapsulate both the information of the node itself and that of its surrounding neighborhood after training. Therefore, we select the output of the \( l \) layer as the semantics of each node. This process can be formally described as follows:
\begin{equation}
\begin{aligned} 
& H_l = GNN_{\{1\sim l\}}\left(A,X,W_{(1,\cdots,l)}^\ast\right) \\ 
& \text{s.t.}  W_{(1,\cdots,L)}^\ast = \arg\min_{W_{(1,\cdots,L)}} \mathcal{L}\left(f_{W_{(1,\cdots,L)}}\left(G\left(A, X\right)\right), Y\right),
\end{aligned}
\label{eq4}
\end{equation}
where \( \mathcal{L}\left(.,.\right) \) is the loss function, commonly the cross-entropy loss. \( W_{(1,\cdots,L)} \) are weight matrices. \(GNN_{\{1\sim l\}} \) refer to the first \( l \) layers of the GNN.

\paragraph{2) Construct adversarial latent representation that preserve the primary semantics:}
\cite{54} address the challenge of imbalanced datasets by extracting the feature shifts of frequent-class entities and applying them to rare-class entities. In this approach, the latent representation is assumed to be \( h = P + T \), where \( P \) denotes the data distribution center (i.e., the similar primary semantics of same-class entities under the fixed task), and \( T \) represents the feature shift (i.e., the secondary semantics of the entities in the task). Inspired by this, we make the following assumption about the composition of the latent representation in our work:
\begin{equation}
h_j^i = dP(i) + eT(j),
\label{eq5}
\end{equation}
where \( h_j^i \) denotes the latent representation vector of the \( j \)-th node in class \( i \). It is composed of two parts: the primary semantics \( dP(i) \), which is crucial for the classification task, and the secondary semantics \( eT(j) \), which is less decisive. All nodes of the same class share the same primary semantic base \( P(i) \), but may have different secondary semantic bases \( T(j) \). Since each class contains some nodes, there exists a set where the primary semantics are preserved while the secondary semantics change. When perturbing the latent representation \( h_j^i \) of each node, we assign a weight \( \alpha_k \) to each representation of the same class. Summing these weighted representations yields the perturbed representation \( \hat{h}_j^i \).
\begin{equation}
\hat{h}_j^i = \frac{\sum_{k=1}^{j} \alpha_k h_k^i}{\sum_{k=1}^{j} \alpha_k}.
\label{eq6}
\end{equation}
Using Equation \eqref{eq5}, the representation \( \hat{h}_j^i \) in Equation \eqref{eq6} can be derived as follows:
\begin{equation}
\begin{aligned} 
\hat{h}_j^i &= \frac{\sum_{k=1}^{j} \alpha_k h_k^i}{\sum_{k=1}^{j} \alpha_k} = \frac{\sum_{k=1}^{j} \alpha_k \left(d_k P(i) + e_k T(k)\right)}{\sum_{k=1}^{j} \alpha_k}\\
&= \frac{\sum_{k=1}^{j} \alpha_k d_k P(i) + \alpha_k e_k T(k)}{\sum_{k=1}^{j} \alpha_k} \\
&= \left(\frac{\sum_{k=1}^{j} \alpha_k d_k}{\sum_{k=1}^{j} \alpha_k}\right) \ast P(i) + \frac{\sum_{k=1}^{j} \alpha_k e_k T(k)}{\sum_{k=1}^{j} \alpha_k}.
\end{aligned}
\label{eq7}
\end{equation}
As derived above, \( \hat{h}_j^i \) still maintains the form \( dP + eT \), where the secondary component is composed of various secondary semantics, and the primary component remains but with a modified coefficient for the primary semantics base. To ensure that the perturbed latent representation \( \widehat{H_l} \) has adversarial effect on the GNN, \( \widehat{H_l} \) is fed into the downstream network layers of the GNN. The optimal \( \alpha \) is then determined based on the loss function \( \text{loss}_1 \). \([h]\) refers to arranging \( h \) in the same order as before.
\begin{equation}
\begin{aligned} 
\mathcal{L}_1 = -\mathcal{L}(Y,GNN_{\{l\sim L\}}\left(A,\widehat{H_l}(\alpha),W_{(l,\cdots,L)}^\ast\right)) \\
- \beta \, \text{sim}(H_l, \widehat{H_l}(\alpha)),
\end{aligned}
\label{eq8}
\end{equation}
where \( \text{sim}(.,.) \) denotes KL divergence. Now we can obtain \( \widehat{H_l}(\alpha) \) as follows:
\begin{equation}
\widehat{H_l}(\alpha) = \left[\hat{h}_j^i\right] = \left[\frac{\sum_{k=1}^{j} \alpha_k h_k^i}{\sum_{k=1}^{j} \alpha_k}\right].
\label{eq9}
\end{equation}

To use a smaller attack budget when reconstructing adversarial samples with perturbed representations, we introduce a penalty term \( \text{sim}(H_l, \widehat{H_l}(\alpha)) \). This term controls the similarity between the perturbed hidden layer and the original hidden layer, thereby limiting the perturbations to a certain range. Notably, it also prevents the coefficient of the primary semantics base from becoming too small. \( \beta > 0 \) is a hyperparameter that controls the strength of the penalty.

\paragraph{3) Map adversarial latent representation to adversarial sample:}To construct an adversarial sample, we need to map the perturbed latent representation to the graph. Let \( \hat{A} \) denote the attacked graph. We introduce a binary perturbation matrix \( S \in \{0,1\}^{N \times N} \) to encode whether an edge in \( G \) is modified. Specifically, an edge \((i, j)\) is modified (added or removed) if \( S_{ij} = S_{ji} = 1 \). Otherwise, if \( S_{ij} = S_{ji} = 0 \), the edge \((i, j)\) remains undisturbed. Given the adjacency matrix \( A \), its complement matrix \( \bar{A} \) is defined as \( \bar{A} = 11^T - I - A \), where \( I \) is the identity matrix, and \( 1 \) is a column vector of all ones. The term \( (11^T - I) \) corresponds to a fully connected graph. Using the edge perturbation matrix \( S \) and the complement matrix \( \bar{A} \), Equation \eqref{eq10} provides the perturbed topology \( \hat{A} \) of the graph \( A \).
\begin{equation}
\hat{A} = A + C \odot S, \quad C = \bar{A} - A,
\label{eq10}
\end{equation}
where \( \odot \) denotes the element-wise product. In the above expression, the positive entries of \( C \) indicate edges that can be added to the graph \( A \), while the negative entries of \( C \) indicate edges that can be removed from \( A \). Due to the difficulty in solving the problem under the binary constraint \(\{0,1\}\), this constraint is relaxed to the continuous interval \([0,1]\). Based on Equation \eqref{eq4}, the latent representation is computed by using the modified graph structure \( \hat{A} \), which is similar to \( \widehat{H_l} \). Hence, we give the mapping loss as follows:
\begin{equation}
\mathcal{L}_2 = -\text{sim}\left(GNN_{\{1\sim l\}}\left(\hat{A},X,W_{(1,\cdots,l)}^\ast\right),\widehat{H_l}\right)
\text{ s.t. } s \in S,
\label{eq11}
\end{equation}
where \( S = \{ s \mid 1^T s \le \epsilon, \; s \in [0,1]^N \} \). The function \( \text{sim}(.,.) \) computes the distance between matrices, commonly using KL divergence. After determining \( s \) based on \( \mathcal{L}_2 \), the values in \( s \) are used to sample from \( A \), resulting in the perturbed graph \( \hat{A} \).

\subsection{Model Optimization}
We solve AHSG step by step. In the first step, we solve for \( W \) by using a conventional gradient descent algorithm. In the second step, when solving the \( \alpha \) problem, the number of elements in \( \alpha \) corresponding to each node is variable because the number of same-class nodes for each node is not fixed. It is challenging to solve for unknowns of variable length in a unified manner. However, the maximum length of \( \alpha \) is fixed, which is the total number of nodes. We set each \( \alpha \) to the maximum length, and after each gradient descent iteration, we perform clipping according to the following formula:

\begin{equation}
\tau (\alpha_{i,j}) =
\begin{cases} 
1, & \text{if } (i = j \text{ and } y_i \text{ is unknown}) \\
\alpha_{i,j}, & \text{if } (y_i = y_j \text{ and } y_i \text{ and } y_j \text{ are known}) \\
0, & \text{else}.
\end{cases}
\label{eq13}
\end{equation}
For the \( s \) subproblem, we scale the hard constraints \(\{0,1\}\) to \([0,1]\). However, under the constraint \( 1^T s \le \epsilon \), we still need to project the gradient-descent result of \( s \) by Equation \eqref{eq14}. For solving \( \mu \), we use the bisection method.
\begin{equation}
\mathrm{\Pi}_S(a) =
\begin{cases}
P_{[0,1]}[a], & \text{if } \sum P_{[0,1]}[a] \le \epsilon \\
P_{[0,1]}[a - \mu], & \text{if } \mu > 0 , \sum P_{[0,1]}[a - \mu] = \epsilon,
\end{cases}
\label{eq14}
\end{equation}
where \( P_{[0,1]}(x) \) is defined as:
\begin{equation}
P_{[0,1]}(x) =
\begin{cases}
x, & \text{if } x \in [0, 1] \\
1, & \text{if } x > 1 \\
0, & \text{if } x < 0.
\end{cases}
\label{eq15}
\end{equation}
Sampling \( A \) using \( s \) produces the attacked graph \( \hat{A} \). The overall procedure of AHSG is shown in Algorithm \ref{alg:algorithm}.

\section{Experiment}
In this section, we validate the effectiveness of AHSG through comprehensive experiments.
\subsection{Experimental Settings}
We evaluated AHSG on three well-known datasets: Cora, Citeseer, and Cora-ML. The statistical results are shown in Table \ref{table1}. 
All datasets contain unweighted edges, allowing the generation of an adjacency matrix \( A \), and include sparse bag-of-words feature vectors that can be used as input for GNN. Among them, Cora and Citeseer are symmetric matrices (i.e., undirected graphs), while Cora-ML is an asymmetric matrix (i.e., directed graph). For this experiment, we converted the Cora-ML dataset into an undirected graph. The Cora and Citeseer datasets have binary features, while the Cora-ML dataset contains continuous features. Following the setup from previous work, the feature vectors of all nodes are fed into the GNN, with only 140 and 120 training nodes for Cora and Citeseer, respectively, and 200 training nodes for the Cora-ML dataset. The number of test nodes is 1,000 for all three datasets.

The baselines include Random, Meta-Self \cite{18}, GradArgmax \cite{16}, PGD \cite{17}, EpoAtk \cite{55} and GraD \cite{20}. We choose GCN as the surrogate model, where the total number of layers \(L\) in the neural network is 2, and \(l\) is 1. We conduct comparative experiments with advanced graph attack and defense methods. For Svd and Jaccard, we adjust the order of high-order approximations and the threshold for removing low-similarity links to achieve optimal defense performance. Other attack methods use default parameters. 

\begin{table}[h!]
\centering
\begin{tabular}{lcccccc}
\toprule
Datasets & Nodes & Links & Features & Classes & Binary \\
\midrule
Cora     & 2708  & 5278  & 1433     & 7       & Y \\
Citeseer & 3327  & 4552  & 3703     & 6       & Y \\
Cora-ML  & 2995  & 8158  & 2879     & 7       & N \\
\bottomrule
\end{tabular}
\caption{Statistics of datasets. The last column indicates whether the dataset has binary features.}
\label{table1}
\end{table}

In our method, the default settings are as follows: the number of iterations \( E \) for perturbation is set to 300, with a learning rate \( \lambda \) of 0.3. The regularization coefficient \( \beta \) is 0.2. For reconstruction, the number of iterations \( T \) is 300, and the dynamic learning rate \( \eta_t \) is \( \frac{q}{\sqrt{t+1}} \), where \( t \) denotes the iteration count and \( q \) is set to 20. The number of selections \( K \) for the probability matrix is 20. When using a two-layer GCN as the victim model, the dimension \( h \) of the first layer is 128, and the second layer’s dimension corresponds to the number of classes.

\begin{algorithm}[tb]
\caption{The procedure of AHSG}
\label{alg:algorithm}
\textbf{Input}: Original graph \( G = (A, X) \)\\
\textbf{Parameter}: \( E \),\( \lambda \),\( \beta \),\( T \),\( \eta_t \),\( K \),\( h \)\\
\textbf{Output}:  \( \hat{A} \)
\begin{algorithmic}[1] 
\STATE Train GNN to obtain $W_{(1,2)}^\ast$.
\STATE Calculate $H_l$ via Equation\eqref{eq4}.
\FOR{$e \gets 1 \text{ to } E$}
    \STATE Calculate $\widehat{H_l}$ via Equation\eqref{eq9}.
    \STATE Gradient descent:\\$\alpha(e) = \alpha(e - 1) - \lambda\nabla \mathcal{L}_1\left( \alpha(e - 1) \right)$.
    \STATE Calculate $\alpha$ via Equation\eqref{eq13}.
\ENDFOR
\FOR{$t \gets 1 \text{ to } T$}
    \STATE Gradient descent:\\$a(t) = s(t-1) - \eta_t \nabla \mathcal{L}_2(s(t-1))$
    \STATE Call projection operation in Equation\eqref{eq14}.
\ENDFOR
\FOR{$k \gets 1 \text{ to } K$}
    \STATE Generate $p$ of the same size as $s$ that follows a uniform distribution.
    \STATE Draw binary vector $d^{(k)}$ following\\ $d_i^{(k)} =
\begin{cases}
1, & \text{if } s_i > p_i \\
0, & \text{if } s_i \leq p_i
\end{cases}$

\ENDFOR
\STATE Choose a vector $s^\ast$ from $d^{(k)}$ which yields the smallest $\text{loss}_2$ under $1^T s \leq \epsilon$.
\STATE Calculate $\hat{A}$ via Equation\eqref{eq10}.
\end{algorithmic}
\end{algorithm}

\subsection{Attack Performance on GNNs}
Except for ProGNN, other GNNs use the evasion attack setting. ProGNN performs both graph structure purification and parameter learning simultaneously, so the attack occurs before training. The attack results on different GNNs by modifying 10\% of the edges are shown in Table \ref{table2}. We observe that all baseline methods lead to a performance drop in the victim models. AHSG outperforms all other attackers on all datasets. Furthermore, AHSG achieves the best attack performance on GCN among different GNN models, since GCN is the corresponding surrogate model, indicating a white-box attack scenario. Furthermore, even against the five defensive GNNs, AHSG still demonstrates strong attack performance, highlighting its robust generalization capability derived from the surrogate model.

\begin{table*}[t]
\centering
\small
\begin{tabular}{llcccccccc}
\toprule
Dataset & Method & Clean & Random & Meta-Self & GradArgmax & PGD & EpoAtk & GraD & AHSG \\ 
\midrule
\multirow{6}{*}{Cora} 
 & GCN      & 0.823 & 0.809 & 0.809 & 0.716 & 0.708 & 0.685 & 0.673 & \textbf{0.646} \\ 
 & Jaccard  & 0.786 & 0.784 & 0.736 & 0.712 & 0.759 & 0.693 & 0.703 & \textbf{0.676} \\ 
 & SVD      & 0.729 & 0.711 & 0.665 & 0.673 & 0.686 & 0.689 & 0.715 & \textbf{0.648} \\ 
 & ProGNN   & 0.809 & 0.794 & 0.659 & 0.737 & 0.703 & 0.673 & 0.665 & \textbf{0.654} \\ 
 & SimPGCN  & 0.790 & 0.782 & 0.784 & 0.703 & 0.728 & 0.703 & 0.695 & \textbf{0.682} \\ 
 & RGCN     & 0.801 & 0.748 & 0.753 & 0.690 & 0.678 & 0.671 & 0.664 & \textbf{0.651} \\ 
\midrule
\multirow{6}{*}{Citeseer} 
 & GCN      & 0.666 & 0.655 & 0.650 & 0.578 & 0.593 & 0.549 & 0.542 & \textbf{0.528} \\ 
 & Jaccard  & 0.665 & 0.632 & 0.645 & 0.617 & 0.623 & 0.591 & 0.622 & \textbf{0.576} \\ 
 & SVD      & 0.601 & 0.593 & 0.573 & 0.588 & 0.565 & 0.589 & 0.574 & \textbf{0.558} \\ 
 & ProGNN   & 0.683 & 0.625 & 0.531 & 0.566 & 0.603 & 0.597 & 0.582 & \textbf{0.530} \\ 
 & SimPGCN  & 0.656 & 0.646 & 0.648 & 0.585 & 0.615 & 0.613 & 0.602 & \textbf{0.581} \\ 
 & RGCN     & 0.610 & 0.579 & 0.569 & 0.576 & 0.573 & 0.559 & 0.552 & \textbf{0.534} \\ 
\midrule
\multirow{6}{*}{Cora-ML} 
 & GCN      & 0.859 & 0.851 & 0.827 & 0.743 & 0.791 & 0.741 & 0.738 & \textbf{0.733} \\ 
 & Jaccard  & 0.860 & 0.839 & 0.807 & 0.770 & 0.828 & 0.801 & 0.813 & \textbf{0.762} \\ 
 & SVD      & 0.820 & 0.799 & 0.794 & 0.759 & 0.816 & 0.763 & 0.808 & \textbf{0.757} \\ 
 & ProGNN   & 0.840 & 0.805 & 0.791 & 0.749 & 0.823 & 0.783 & 0.769 & \textbf{0.735} \\ 
 & SimPGCN  & 0.846 & 0.836 & 0.817 & 0.743 & 0.818 & 0.774 & 0.758 & \textbf{0.740} \\ 
 & RGCN     & 0.859 & 0.842 & 0.831 & 0.754 & 0.828 & 0.785 & 0.771 & \textbf{0.755} \\ 
\bottomrule
\end{tabular}
\caption{Accuracy of GNNs with 10\% edge modifications. The best result in each row is highlighted in bold.}
\label{table2}
\end{table*}

\begin{table*}[t]
\centering
\small
\begin{tabular}{lcccccccc}
\toprule
Dataset & Attack Ratio & Random & Meta-Self & GradArgmax & PGD & EpoAtk & GraD & AHSG \\ 
\midrule
\multirow{4}{*}{Cora} 
 & 5\%  & 0.814 & 0.814 & 0.765 & 0.756 & 0.731 & 0.725 & \textbf{0.698} \\ 
 & 10\% & 0.809 & 0.809 & 0.716 & 0.708 & 0.685 & 0.673 & \textbf{0.646} \\ 
 & 15\% & 0.795 & 0.805 & 0.685 & 0.681 & 0.634 & \textit{OOM} & \textbf{0.615} \\ 
 & 20\% & 0.787 & 0.798 & 0.652 & 0.645 & 0.607 & \textit{OOM} & \textbf{0.585} \\ 
\midrule
\multirow{4}{*}{Citeseer} 
 & 5\%  & 0.657 & 0.651 & 0.608 & 0.610 & 0.585 & 0.578 & \textbf{0.565} \\ 
 & 10\% & 0.655 & 0.650 & 0.578 & 0.593 & 0.549 & 0.542 & \textbf{0.528} \\ 
 & 15\% & 0.653 & 0.631 & 0.539 & 0.564 & 0.523 & \textit{OOM} & \textbf{0.495} \\ 
 & 20\% & 0.641 & 0.628 & 0.507 & 0.529 & 0.501 & \textit{OOM} & \textbf{0.487} \\ 
\midrule
\multirow{4}{*}{Cora-ML} 
 & 5\%  & 0.853 & 0.841 & 0.791 & 0.843 & 0.795 & 0.789 & \textbf{0.776} \\ 
 & 10\% & 0.851 & 0.827 & 0.743 & 0.791 & 0.741 & 0.738 & \textbf{0.733} \\ 
 & 15\% & 0.846 & 0.821 & 0.721 & 0.751 & 0.711 & \textit{OOM} & \textbf{0.700} \\ 
 & 20\% & 0.837 & 0.809 & 0.695 & 0.708 & 0.698 & \textit{OOM} & \textbf{0.685} \\ 
\bottomrule
\end{tabular}
\caption{Accuracy of GCN under different attack budgets. The best result in each row is highlighted in bold. OOM represents Out of Memory.}
\label{table3}
\end{table*}

\subsection{Attack Performance w.r.t Attack Budget}
Table \ref{table3} presents the accuracy on the test sets on three datasets as the proportion of perturbed edges increases from 5\% to 20\%. For example, in the case of the Cora dataset, as anticipated, the test accuracy consistently decreases with a higher number of perturbed edges, although the rate of decline slows down. Moreover, the proposed AHSG achieves the best attack performance across all perturbation ratios. Similar conclusions are drawn for the Citeseer and Cora-ML datasets.

\subsection{Semantic Detection}
In the fields of computer vision (CV) and natural language processing (NLP), humans can effectively perform semantic checks. However, in the domain of graph, it is challenging for humans to inspect the semantics of large-scale graphs. To address this, \cite{53} attempt to define semantic boundaries by introducing a reference classifier \( g \) to represent changes in semantic content. The reference classifier \( g \) can be derived from knowledge about the data generation process. According to the data generation process of CSBMs, \cite{53} use a Bayes classifier as \( g \). 

Data generation process: synthetic graphs with analytically tractable distributions are generated using the Contextual Stochastic Block Models (CSBMs). It defines the edge probability \( p \) between same-class nodes and \( q \) between different-class nodes. Node features are extracted using a Gaussian model. Sampling from CSBMs can be described as an iterative process for node \( i \in {n} \):
\begin{enumerate}
    \item Sample the label \( y_i \sim \text{Ber}(1/2) \) (Bernoulli distribution).
    \item Sample the feature vector \( x_i \mid y_i \sim \mathcal{N}((2y_i-1)\mu, \sigma I) \), where \( \mu \in \mathbb{R}^d \) and \( \sigma \in \mathbb{R} \).
    \item For all \( j \in {n} \), if \( y_i = y_j \), then sample \( A_{j,i} \sim \text{Ber}(p) \); otherwise, sample \( A_{j,i} \sim \text{Ber}(q) \) and set \( A_{i,j} = A_{j,i} \).
\end{enumerate}
We denote this as \( (X, A, y) \sim \text{CSBM}(n, p, q, \mu, \sigma^2) \).

We reclassify the nodes in the attacked graph \( G(X', A') \), which is generated by Equation \eqref{eq17}.
\begin{equation}
G(X', A') = \text{attack}(G(X, A)). 
\label{eq17}
\end{equation}
As shown in Equation \eqref{eq16}, if the reference classifier \( g \) gives the same prediction for node \( v \) in both the original and attacked graph, and the prediction is correct, then \cite{53} consider that the primary semantics of node \( v \) have not been altered by the attack.
\begin{equation}
\begin{aligned} 
g(X', A')_v = g(X, A)_v = y_v, \\
(X, A, y) \sim \text{CSBM}(n, p, q, \mu, \sigma^2). 
\end{aligned}
\label{eq16}
\end{equation}
In Equation \eqref{eq19}, the reference classifier \( g \) consists of a feature-based Bayesian classifier and a structure-based Bayesian classifier. The structure-based Bayesian classifier considers four scenarios: whether nodes of the same class are connected and whether nodes of different classes are connected.
\begin{equation}
\begin{aligned} 
g(X,A)_v = \arg\max_{y} \left( Bayes(X, A, v, y) \right),\\
Bayes(X,A,v,y) = log( p\left( X_{v} \middle| y_{v} \right) +\\ log( {\prod\limits_{i = 0}^{n}p^{A\lbrack i,v\rbrack(1 - |y_{i} - y_{v}|)}}(1 - p)^{(1 - A\lbrack i,v\rbrack)(1 - |y_{i} - y_{v}|)}\\q^{A\lbrack i,v\rbrack(|y_{i} - y_{v}|)}(1 - q)^{(1 - A\lbrack i,v\rbrack)(|y_{i} - y_{v}|)} ).
\end{aligned}
\label{eq19}
\end{equation}
The number of nodes with consistent classification results before and after the attack can be regarded as the degree of graph semantics preservation. We denote this by \( \text{Bayes\_maintain} \), as shown in Equation \eqref{eq20}, where \( \mathbb{I} \) is an indicator function. 
\begin{equation}
\text{Beyes\_maintain} = \frac{\sum_{i=0}^n \mathbb{I}_{g(X', A')_v = g(X, A)_v = y_v}}{n}.
\label{eq20}
\end{equation}
As shown in Table \ref{table4}, the Bayes reference classifier achieves 92.4\% accuracy on clean graphs, validating its effectiveness as a semantic proxy. Comparison with other attack methods reveals that AHSG alters the Bayesian classification results the least, thereby maintaining the majority of the primary semantics in the classification task while achieving the best attack performance.

\begin{table}[!ht]
\centering
\small
\begin{tabular}{lcc}
\toprule
Method      & GCN   & Beyes\_maintain \\ 
\midrule
Clean       & 0.724 & 0.924 \\ 
Meta-Self   & 0.653 & 0.810 \\ 
GradArgmax  & 0.632 & 0.802 \\ 
PGD         & 0.607 & 0.832 \\ 
EpoAtk      & 0.597 & 0.774 \\ 
GraD        & 0.593 & 0.783 \\ 
AHSG        & 0.586 & 0.903 \\ 
\bottomrule
\end{tabular}
\caption{Accuracy of GCN and the degree of primary semantic preservation under different attack methods.}
\label{table4}
\end{table}

Note that in CSBMs, We use the original default settings. The number of generated nodes \( n \) is 1000. Node feature mean \( \mu \) is calculated as \( \frac{M\sigma}{2\sqrt{d}} \), with variance \( \sigma = 1 \), \( d = \frac{n}{(\ln(n))^2} = 21 \), and \( M = 0.5 \). Smaller value of \( M \) causes the reference classifier to rely more on structural information. Since AHSG focuses on structural attacks, it is essential for the reference classifier to analyze the semantics within the structure. The edge probability \( p \) between same-class nodes is 0.6326\%, while the edge probability \( q \) between different-class nodes is 0.1481\%.

\subsection{Ablation Study}
To further validate the effectiveness of AHSG, we conduct an ablation study to analyze the impact of its various components. Since AHSG consists of a perturbation module, a reconstruction module, and a regularization module, we performed ablation study by removing one of these modules to assess their influence on AHSG's performance. The regularization module is discussed in the hyperparameter analysis. The specific combinations for the ablation study are as follows: AHSG-rec refers to the version where we retain the perturbation module of AHSG but choose random connections during the reconstruction process. AHSG-hid refers to the version where we retain the reconstruction module of AHSG but apply random perturbations during the perturbation process. As shown in Table \ref{table5}, the best attack performance is achieved only when both perturbation and reconstruction steps are present simultaneously. This result is reasonable. Without the guidance of the latent representation, the reconstruction step cannot capture the correct perturbation information. Conversely, without the reconstruction step, the perturbed latent representation cannot be transformed into an effective attack graph.
\begin{table}[h!]
\centering
\small
\begin{tabular}{lccc}
\toprule
Modules     & Cora   & Citeseer & Cora-ML \\ 
\midrule
Clean       & 0.823  & 0.666    & 0.859   \\ 
AHSG-rec    & 0.809  & 0.655    & 0.852   \\ 
AHSG-hid    & 0.804  & 0.650    & 0.842   \\ 
AHSG        & 0.698  & 0.565    & 0.776   \\ 
\bottomrule
\end{tabular}
\caption{Accuracy of GCN with 10\% edge modifications on different modules and datasets.}
\label{table5}
\end{table}

\begin{figure}[htbp]
 \centering
    \subfigure{
        \includegraphics[width=0.2\textwidth]{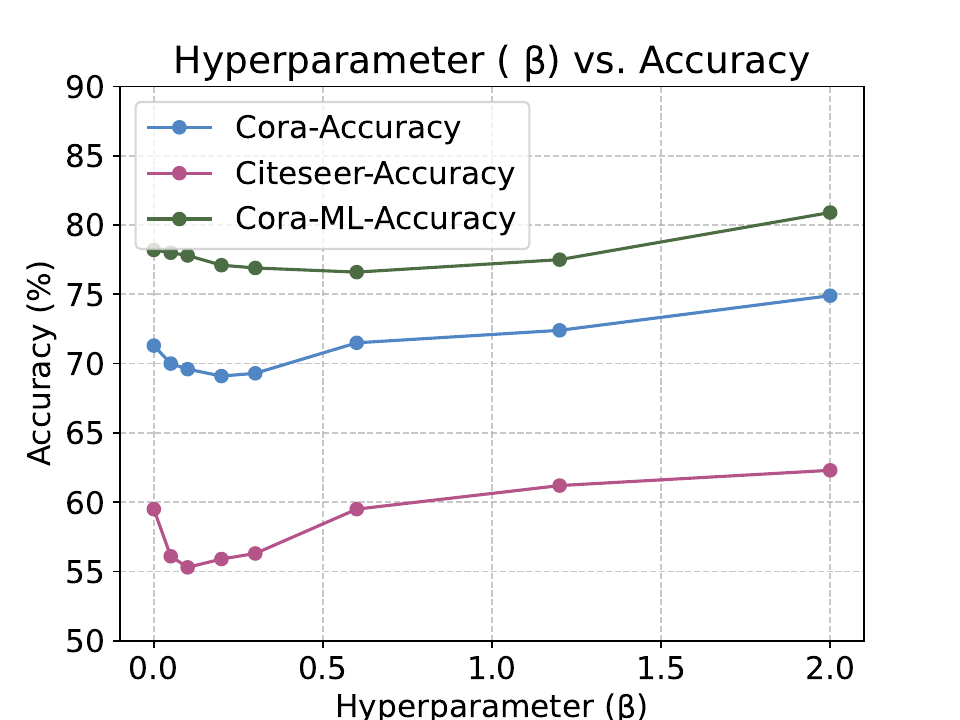}    
    }
    \subfigure{
        \includegraphics[width=0.2\textwidth]{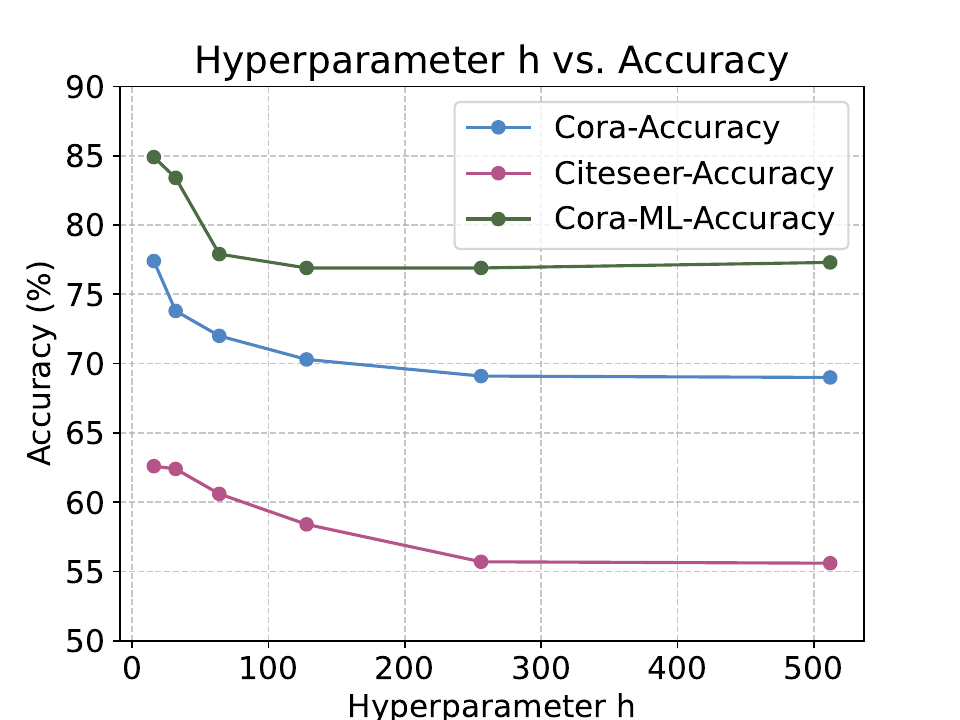}     
    }
\caption{Hyperparameter analysis on the regularization term coefficient \( \beta \) and the hidden layer dimension \( h \). }
\label{fig3}
\end{figure}
\subsection{Parameter Analysis}
Finally, we performed a hyperparameter analysis on the regularization term coefficient \( \beta \) and the hidden layer dimension \( h \). 
AHSG achieves optimal performance when \( \beta \) is around 0.1, as this value effectively balances the contributions of the attack loss and the regularization term. Large \( \beta \) restricts the range of perturbation in the hidden layers, preventing the attack from fully utilizing the attack budget. Conversely, small \( \beta \) can result in an excessively large distance between the perturbed representation and the original representation, making it difficult to reconstruct the perturbed representation within a given attack budget. When changing the hidden layer dimension \( h \) of the surrogate model, AHSG's performance shows considerable fluctuation. Particularly, when the hidden layer dimension is small, the attack effectiveness decreases significantly. Because the hidden layers with a small dimension can not adequately capture the information of each node (including both primary and secondary semantics) for AHSG to utilize. However, it is worth noting that an attack graph generated using a surrogate model with a large hidden layer dimension can still be effective when applied to a target model with a small hidden layer dimension.

\section{Conclusion}
We investigate graph structure attacks on GNNs under the evasion setting and propose AHSG, which preserves primary semantics in hidden layers to prevent significant semantic disruption. By exploiting the similarity of primary semantics among nodes with the same class, AHSG constructs representations that cause GNN failure while maintaining semantic consistency. Adversarial graphs are generated using the PGD algorithm, with a regularization term added to ensure semantic invariance. Experiments on various datasets show AHSG's superior attack performance against multiple GNNs, including defense models. Semantic detection confirms that AHSG effectively preserves task-relevant primary semantics.

\bibliographystyle{named}
\bibliography{ijcai25}
\end{document}